\pdfoutput=1

\documentclass[11pt]{article}

\usepackage[final]{acl}

\usepackage{times}
\usepackage{latexsym}

\usepackage[T1]{fontenc}

\usepackage[utf8]{inputenc}

\usepackage{microtype}

\usepackage{inconsolata}

\usepackage{graphicx}

\usepackage{amsmath}
\usepackage{amsfonts}
\usepackage{booktabs}
\usepackage{multirow}

\title{Personality Vector: Modulating Personality of Large Language Models by Model Merging}

\author{
    \textbf{Seungjong Sun}$^{*1}$, \textbf{Seo Yeon Baek}$^{*2}$, \textbf{Jang Hyun Kim}$^{\dagger1,2}$ \\
    $^{1}$Department of Human-Artificial Intelligence Interaction, Sungkyunkwan University \\
    $^{2}$Department of Immersive Media Engineering, Sungkyunkwan University \\ 
    \texttt{\{tmdwhd406, qortjdus1999\}}@g.skku.edu, \\ \texttt{ alohakim}@skku.edu}

\begin{document}
\maketitle
\renewcommand{\thefootnote}{\fnsymbol{footnote}} 
\footnotetext[1]{Equally contributed}
\footnotetext[2]{Corresponding author}
\renewcommand{\thefootnote}{\arabic{footnote}}
\begin{abstract}
Driven by the demand for personalized AI systems, there is growing interest in aligning the behavior of large language models (LLMs) with human traits such as personality. Previous attempts to induce personality in LLMs have shown promising results, but they struggle to capture the continuous and multidimensional nature of human traits. In this work, we propose a novel method for personality modulation in LLMs via model merging. Specifically, we construct personality vectors by subtracting the weights of a pre-trained model from those of the  fine-tuned model on a given personality trait. By merging personality vectors, we enable LLMs to exhibit desired personality traits without additional training. Extensive experiments show that personality vectors enable continuous control over trait intensity and support the composition of multiple traits. Furthermore, personality vectors transfer across diverse downstream models, suggesting that they encode generalizable representations of personality. Our code is available at \href{https://github.com/RSS-researcher/Personality_vector.git}{here}.

\end{abstract}

\section{Introduction} 
Large Language Models (LLMs) have not only demonstrated human-like language capabilities but are also increasingly found to exhibit behaviors aligned with human cognitive and psychological traits \citep{bai2022training}. As the demand for personalized AI agents grows, recent research has explored ways to modulate LLM behavior based on individual personality characteristics \citep{jang2023personalized,tseng2024two, zhu2024personality}. Among these, aligning model outputs with established personality frameworks such as the Big Five personality traits \citep{mccrae1987validation} has emerged as a promising direction for developing more human-aligned and personalized LLM.

Previous work on personality induction in LLMs has explored various approaches, including prompt-based methods \citep{jiang2023evaluating, serapio2023personality, jiang2024personallm}, fine-tuning \citep{pan2023llms, chen2024Extroversion, cui2023machine}, and activation intervention \citep{zhu2024personality, li2023tailoring, deng2024neuron, weng2024controllm}. While existing approaches have guided model behavior toward specific personality types, little work has addressed fine-grained personality control in LLMs. Since every individual has a unique personality — varying in both type and intensity \citep{mccrae1987validation, goldberg1992development} — more continuous and multidimensional personality control is crucial for developing advanced personalized AI systems.

\begin{figure}
        \centering
        \includegraphics[width=\columnwidth]{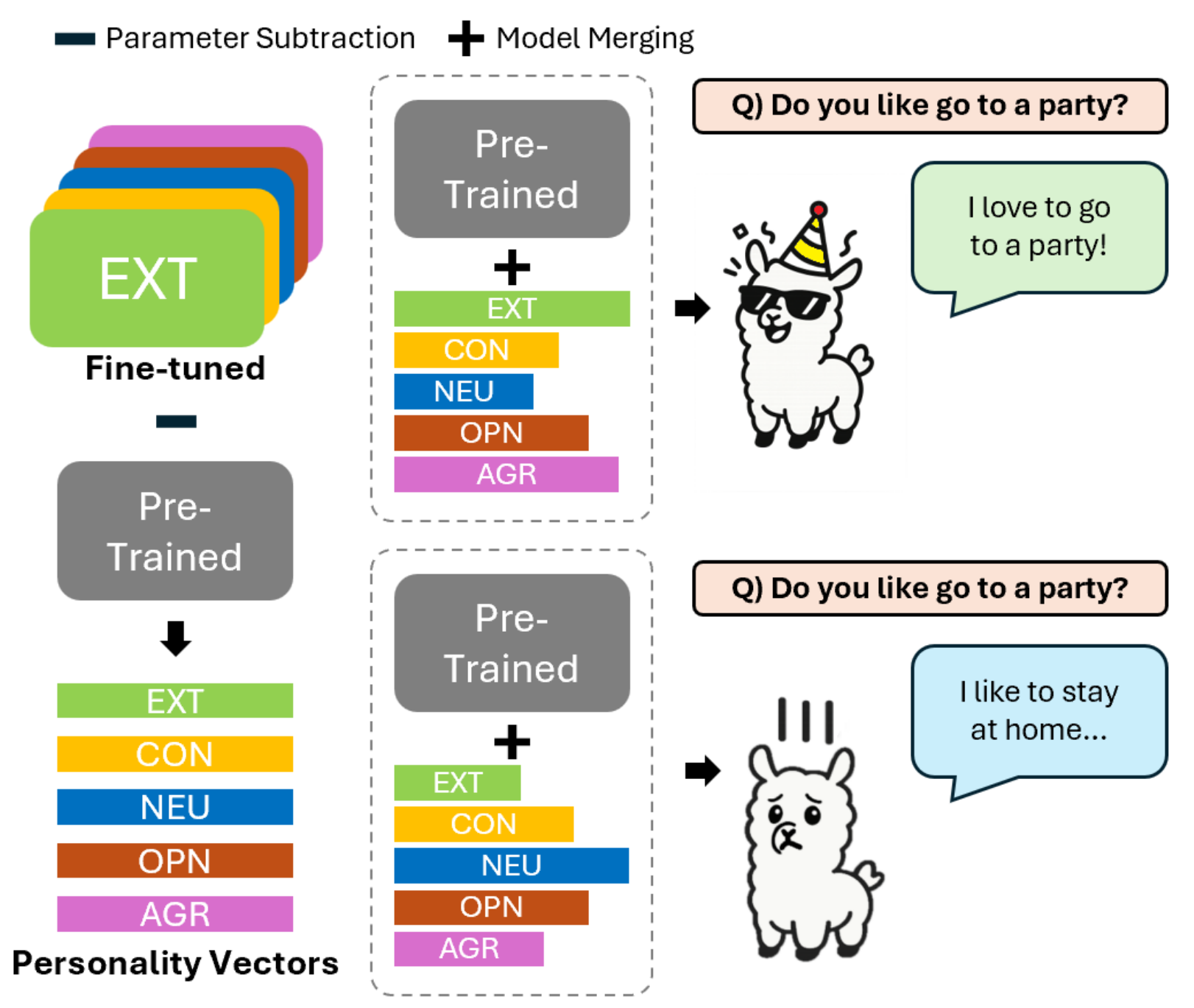}
        \caption{Personality modulation pipeline. For each Big Five personality trait, a personality vector is obtained by subtracting the parameters of the pre-trained model from those of the corresponding fine-tuned model. These vectors can then be merged into other models with a scaling coefficient to control the intensity of personality expression.}
        \label{fig:figure_1}
\end{figure}

To address this gap, we introduce a new approach based on model merging, which enables capability transfer across models via weight-space interpolation \citep{matena2022merging, wortsman2022model, jin2022dataless}. Inspired by the concept of task vectors \citep{ilharco2022editing}, we investigate whether personality traits acquired through fine-tuning can be transferred across models.  As shown in \autoref{fig:figure_1}, we fine-tune models on each of the Big Five personality traits and compute personality vectors by subtracting the weights of the pre-trained base model. These vectors are then merged into other models to induce the desired personality characteristics, enabling personality modulation without additional training.

Our evaluation involved four main experiments: 
(1) We examined whether personality trait intensity is scalable by adjusting the magnitude of the personality vector during merging. (2) We tested whether multiple traits can be integrated simultaneously by merging all five personality vectors into a single model. (3) We evaluated whether subtracting a personality vector from the base model induces the opposite trait. (4) We assessed the transferability of personality vectors to models from different domains, including Role-Playing Agents (RPAs), Korean Language Model, and Vision Language Model (VLM).
Through extensive experiments, our work offers the following contributions:
\begin{itemize}
    \item We propose a model merging based approach to modulate personality without additional training.
    \item We demonstrate fine-grained personality control, including continuous scaling and multi-trait composition.
    \item We validate the transferability of personality vectors across diverse downstream models.
\end{itemize}

\section{Related Work}

\subsection{Personality Assignment in LLMs} 
Prior research has focused on aligning LLM outputs with human personality typologies, such as the Big Five personality \citep{mccrae1987validation} or Myers-Briggs Type Indicator (MBTI) \citep{boyle1995myers}. Existing approaches include prompt-based methods \citep{jiang2024personallm, serapio2023personality, jiang2023evaluating}, which are lightweight but highly sensitive to prompt and lack consistency in long contexts \citep{wang2024investigating}. Training-based methods \citep{pan2023llms, chen2024Extroversion, cui2023machine, li2024big5} offer greater control but require substantial computational resources. More recently, activation intervention techniques have been explored, which identify neurons whose activations vary by trait and manipulate them during inference to steer the model’s responses \citep{meng2022locating, zhu2024personality, deng2024neuron}. Existing research has demonstrated that personality steering in LLM is feasible; however, it struggles to capture the continuous, multi-dimensional nature of human personality \citep{costa1992four, soto2018big}. Some studies have attempted to control trait intensity through neuron-level scaling using activation intervention; however, the results have been limited \citep{li2023tailoring, deng2024neuron, weng2024controllm}. Moreover, multi-trait control has been limited to combining at most two to four traits, falling short of the Big Five personality \citep{li2023tailoring, deng2024neuron}. To address these gaps, our study introduces a model merging-based approach that enables continuous control and simultaneous composition of multiple personality traits.

\subsection{Model Merging} 
Model merging combines knowledge from multiple models via parameter-wise weight interpolation without requiring additional gradient-based training \citep{matena2022merging, wortsman2022model, jin2022dataless, li2022branch}. Building on this idea, task arithmetic \citep{ilharco2022editing} has demonstrated that model abilities can be transferred through arithmetic operations on task vectors, which are defined as the difference between the weights of a fine-tuned model and its corresponding pre-trained base. To enable effective merging of multiple task vectors, recent studies have proposed methods such as TIES-Merging \citep{yadav2023ties} and DaRE (Drop and REscale) \citep{yu2024language} to mitigate parameter interference and preserve task-specific information \citep{hagos2024recent, goddard2024arcee, akiba2025evolutionary}. Initial work on model merging primarily focused on combining task performance, such as reasoning and inference accuracy \citep{huang2024emr, yang2024model, liu2024bitdelta}. More recent studies, however, have begun to explore human-like behavioral modulation—such as controlling chat styles for human alignment \citep{huang2024chat} or adjusting emotional tone in text-to-speech models \citep{kalyan2024emotion}. Inspired by these efforts, we investigate whether personality traits can be transferred across models using model merging techniques.

\section{Methods} 
We explore personality induction in LLMs via personality vector merging, extracting trait-specific vectors from fine-tuned models and integrating them into pre-trained or downstream models.

\subsection{Personality Vector} 
We fine-tuned a pre-trained model $\theta_{pre} \in \mathbb{R}^d$ using personality-specific dialogue datasets to obtain personality vectors based on the Big Five personality traits. For each personality condition $p \in P = \{\text{OPN}_{\text{high}}, \text{OPN}_{\text{low}}, \dots, \text{NEU}_{\text{high}}, \text{NEU}_{\text{low}}\}$, the resulting fine-tuned model has parameters $\theta_{p} \in \mathbb{R}^d$. We define the personality vector as:
\begin{equation}
  \label{eq:personality_vector}
  \phi_{p} = \theta_{p} - \theta_{pre}
\end{equation}
These vectors represent personality-specific task vectors and can be merged into target model $\theta \in \mathbb{R}^d$. We empirically evaluate whether injecting $\phi_{p}$ into a model modifies its output to reflect the associated personality trait.

\subsection{Model Merging} 
\paragraph{Task Arithmetic} \citep{ilharco2022editing} injects or negates capabilities by adding or subtracting a task vector: $\theta' = \theta_{base} + \alpha \phi$, where $\alpha$ is an optional scaling coefficient ($\alpha = 1$ recovers the fully fine-tuned model). We apply this formulation to personality vectors as $\theta' = \theta_{base} + \alpha \phi_{p}$ to evaluate whether personality attributes can be linearly composed into models.

\paragraph{TIES-Merging} \citep{yadav2023ties} addresses parameter interference arising from combining multiple task vectors. It reduces information loss by zeroing out minor parameter updates, aligning signs across task vectors, and merging only parameters with consistent directional changes. We adopt TIES-Merging when integrating multiple personality vectors to reduce parameter interference and preserve salient personality features.

\paragraph{DaRE} \citep{yu2024language} sparsifies task vectors by randomly dropping and rescaling parameters, thereby reducing parameter interference during merging. It samples a random mask $m_k \sim \text{Bernoulli}(p)$ and generates a sparsified vector $\tau^{k} =  \left( (1 - m_k) \odot \tau_k \right)/(1 - p)$, where $\odot$ denotes element-wise multiplication. We apply DaRE alongside task arithmetic and TIES-Merging to mitigate interference when merging multiple personality vectors.

\section{Experimental Setting}
In this section, we describe the validation experiments designed to address the following research questions (RQs):
\begin{itemize}
    \item RQ1: Can a model’s personality intensity be controlled by scaling personality vectors?
    \item RQ2: Can multiple personality traits be combined into a model through vector merging?
    \item RQ3: Can subtracting a personality vector induce opposing traits?
    \item RQ4: Can personality vectors effectively transfer personality traits to  models fine-tuned for different domains?

\end{itemize}

\subsection{Data}
We adopted the Big5-Chat dataset \citep{li2024big5}, which is a dialogue-based dataset constructed around the Big Five personality framework. It contains 100,000 dialogue examples, with 10,000 examples for each of the 10 personality categories. These categories represent high or low levels of the five personality traits: Openness, Conscientiousness, Extraversion, Agreeableness, and Neuroticism. For each trait, we fine-tuned a separate model, resulting in 10 fine-tuned personality-specific models.

\subsection{Models and Baselines}
\paragraph{Models} 
Experiments were conducted using Llama-3.1-8B-Instruct \citep{grattafiori2024llama} and Qwen2.5-7B-Instruct \citep{yang2024qwen2}. Following prior findings that Supervised Fine-Tuning (SFT) outperforms Direct
Preference Optimization (DPO) for imparting personality traits \citep{li2024big5}, we trained each model on the Big5-Chat dataset using SFT. The hyperparameters used for fine-tuning are detailed in Appendix~\ref{sec:appendix-A.1}. We obtained 10 personality vectors by subtracting the base model weights from each fine‑tuned model. All experiments were repeated five times with a temperature of 0.6. We compare our approach against the following baselines: 

\paragraph{Prompt} Prompt-based personality conditioning modifies model behavior by injecting trait descriptive adjectives into the prompt \citep{goldberg1992development, serapio2023personality}. For each trait, five adjectives corresponding to the desired polarity (e.g., High or Low Extraversion) are randomly selected and combined with intensity modifiers such as "very" (high), "a bit" (low), or none (moderate). Full prompt templates are provided in Appendix~\ref{sec:appendix-A.2}. Additionally, we used Personality Prompting (P\textsuperscript{2}) generated by ChatGPT that describe each trait \citep{jiang2023evaluating}; the corresponding results are reported in Figure \ref{fig:figure_9},\ref{fig:figure_10_p2_liwc} and Table \ref{tab:table_9_p2}. Furthermore, the experimental results of GPT-4o model are provided in Appendix~\ref{sec:appendix-A.2}.

\paragraph{NPTI} Neuron-level Personality Trait Intervention (NPTI) steers model behavior by adjusting neuron activations related to specific traits \citep{deng2024neuron}. Trait expression is modulated by amplifying or suppressing relevant neurons using a scaling gamma $\gamma \in [0.1, 2.0]$. Further details are provided in Appendix~\ref{sec:appendix-A.2}.

\subsection{Evaluation}
We applied two complementary tasks to evaluate the personality expressed by the merged models: the Big Five Inventory (BFI) questionnaire and analysis of linguistic features in  generated text.

\paragraph{BFI} The BFI consists of 44 items designed to assess the five major personality traits \citep{john1999big}. It has also been adopted to evaluate personality expression in LLMs \citep{jiang2024personallm, wang2024incharacter}. However, due to a lack of self-awareness, LLMs often struggle to provide reliable responses in self-report questionnaires \citep{cui2023machine, dan2024p}. To address this, we adopted an interview-style format \citep{wang2024incharacter}, where the model responded in natural dialogue, and used the GPT API to score responses on a 5-point scale.  To validate GPT-based evaluation, we compared it with human judgements. Further details are provided in Appendix~\ref{sec:appendix-A.3}.

\paragraph{Linguistic feature} Personality is known to be prominently reflected in language use \citep{norman1963toward, raad2002big, mehl2006personality}. Therefore, we analyzed the linguistic features of model-generated text to evaluate personality expression. Each model was prompted with the instruction: \textit{"Tell me about yourself in 300 words."} \citep{jiang2024personallm} The responses were analyzed using \textsc{LIWC-22}. We constructed trait-specific linguistic features from LIWC outputs. Further details are provided in Appendix~\ref{sec:appendix-A.3}.

\begin{table}[t]
\centering
\resizebox{0.5\textwidth}{!}{%
\begin{tabular}{lccccc}
\toprule
\textbf{} & \textbf{OPN} & \textbf{CON} & \textbf{EXT} & \textbf{AGR} & \textbf{NEU} \\
\midrule
\text{High} & 5.0~$\uparrow$ & 4.89~$\uparrow$ & 4.48~$\uparrow$ & 4.69~$\uparrow$ & 4.38~$\uparrow$ \\
\text{Base} & 4.24 & 3.65 & 2.80 & 4.24 & 2.57 \\
\text{Low} & 2.06~$\downarrow$ & 2.02~$\downarrow$ & 1.95~$\downarrow$ & 1.33~$\downarrow$ & 2.20~$\downarrow$ \\
\bottomrule
\end{tabular}%
}
\caption{BFI personality test results for fine-tuned models. High and Low refer to high-trait and low-trait personality conditions (e.g., High Openness vs. Low Openness). Base indicates the pre-trained model (Llama-3.1-8B-Instruct).}
\label{tab:table_1}
\end{table}

\begin{figure*}[t]
  \centering
  \includegraphics[width=\textwidth]{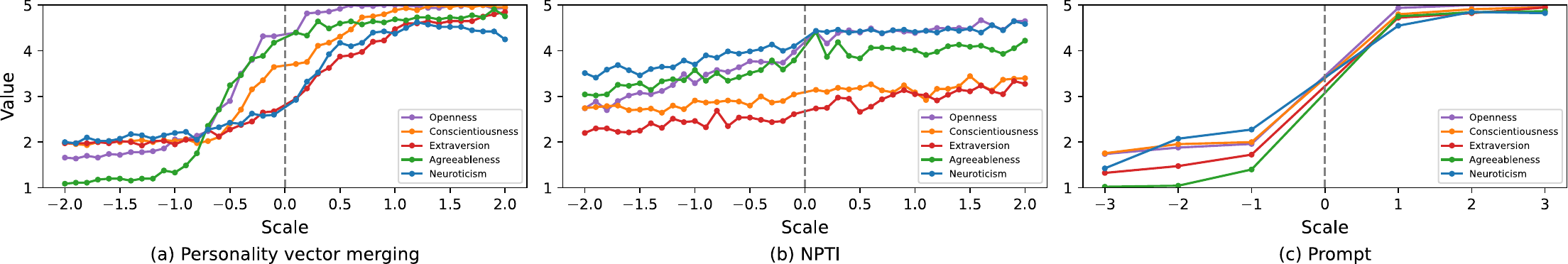}
  \caption{BFI scores across different scaling levels for a single personality trait. Results to the right of 0 represent high-trait conditions; those to the left represent low-trait conditions. (a) Personality vector merging and (b) NPTI were scaled from 0.1 to 2.0, while (c) prompt-based scaling ranged from 1 to 3.}
  \label{fig:figure_2}
\end{figure*}

\begin{figure*}[t]
  \centering
  \includegraphics[width=\textwidth]{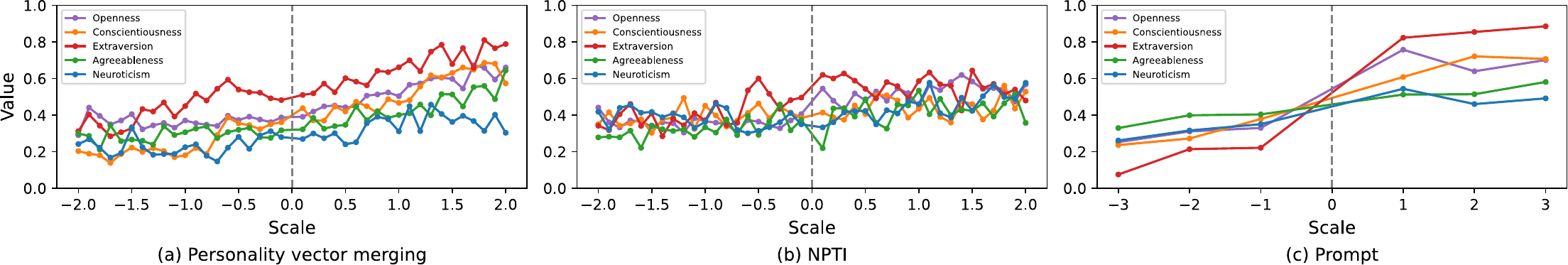}
  \caption{Linguistic feature scores across different scaling levels for a single personality trait. Results to the right of 0 represent high-trait conditions; those to the left represent low-trait conditions. (a) Personality vector merging and (b) NPTI were scaled from 0.1 to 2.0, while (c) prompt-based scaling ranged from 1 to 3.}
  \label{fig:figure_3}
\end{figure*}

\begin{table*}[t]
\centering
\resizebox{\textwidth}{!}{%
\begin{tabular}{lcccccc}
\toprule
\textbf{} & \textbf{Openness} & \textbf{Conscientiousness} & \textbf{Extroversion} & \textbf{Agreeableness} & \textbf{Neuroticism} & \textbf{AVG} \\
\midrule
\textbf{\textit{BFI score}} \\
Prompt & 0.536 & 0.877 & 0.962 & 0.910 & 0.883 & \textbf{0.834} \\
NPTI & 0.729 & 0.302 & 0.487 & 0.691 & -0.065 & 0.429 \\
\midrule
Task arithmetic & 0.548 & 0.694 & 0.686 & 0.525 & 0.420 & 0.575 \\
Task arithmetic + DaRE & 0.547 & 0.772 & 0.712 & 0.641 & 0.560 & 0.646 \\
TIES-Merging & 0.515 & 0.703 & 0.678 & 0.542 & 0.448 & 0.577 \\
TIES-Merging + DaRE & 0.558 & 0.730 & 0.699 & 0.593 & 0.535 & 0.623 \\
\midrule
\midrule
\textbf{\textit{Linguistic feature}} \\
Prompt & 0.253 & 0.268 & 0.541 & 0.208 & 0.174 & 0.289 \\
NPTI & 0.402 & 0.200 & 0.129 & 0.160 & 0.155 & 0.209 \\
\midrule
Task arithmetic & 0.161 & 0.203 & 0.262 & 0.199 & 0.077 & 0.180 \\
Task arithmetic + DaRE & 0.344 & 0.378 & 0.323 & 0.311 & 0.165 & \textbf{0.304} \\
TIES-Merging & 0.191 & 0.273 & 0.231 & 0.255 & 0.094 & 0.209 \\
TIES-Merging + DaRE & 0.298 & 0.441 & 0.279 & 0.230 & 0.163 & 0.282 \\
\bottomrule
\end{tabular}%
}
\caption{Pearson correlations between personality scales and BFI scores (top), and between personality scales and linguistic features (bottom), under the multi-trait merging setting. AVG denotes the average correlation across all five traits. Task arithmetic, TIES-Merging, and DaRE refer to the merging methods used for personality vector.}
\label{tab:table_2}
\end{table*}

\subsection{Main Experiments}
We conducted a series of experiments to examine whether the personality of a base model could be modulated by merging personality vectors in various ways.

\subsubsection{Personality Scaling}
We tested whether each trait \textit{intensity} could be modulated using the scaling coefficient $\alpha$ in $\theta_p^\alpha = \theta_{base} + \alpha \phi_p$, where $p$ denotes one of the 10 high/low personality variants and $\alpha \in [0.1, 2.0]$. For each of the 10 personality vectors, we generated 20 scaled variants, yielding 200 merged models in total.

We evaluated instruction-following performance using \textsc{AlpacaEval} to ensure that the model’s instruction-following capability was preserved \citep{li2023alpacaeval, polo2024tinybenchmarks}. \autoref{fig:figure_11_alpaca}a indicate that model performance remains stable across scales.

\subsubsection{Multi-Personality Composition}
We explored whether an LLM’s personality could be simultaneously modulated by merging multiple personality vectors. We define merged models as: $\theta_{multi}^{\alpha} = \theta_{base} +  \sum_{p \in P} \alpha\phi_p$, with $P = \{\text{OPN}_{\text{high}}, \text{OPN}_{\text{low}}, \dots, \text{NEU}_{\text{high}}, \text{NEU}_{\text{low}}\}$, where $\alpha$ is the scaling coefficient of each personality vector.

We constructed 32 multi-personality models for each value of the scaling coefficient. \textsc{AlpacaEval} results revealed a drop in instruction-following performance when the sum of $\alpha$ exceeded 2.0 (see \autoref{fig:figure_11_alpaca}b). Therefore, we constrained $\alpha$ to the range $[0.1, 0.4]$, yielding 128 models in total.

We applied several merging strategies, such as TIES-Merging, task arithmetic with DaRE, and TIES-Merging with DaRE, to address potential parameter interference from merging multiple vectors. For these experiments, we set the DaRE drop rate to 0.5 and TIES-Merging trimming rate to 0.7. Full details on tuning and optimal merging coefficients are presented in Appendix~\ref{sec:appendix-B.1}.

\subsubsection{Personality Negation}
We tested whether the opposite trait could be induced by subtracting a personality vector. We applied the equation $\theta_{\alpha, p} = \theta_{base} + \alpha \phi_{p}$, with $\alpha = -1$, and subtracted the personality vector from the base model. This allowed us to assess whether the traits could be directionally reversed through negative scaling.

\subsection{Transferability}
To assess the extensibility of personality vectors, we tested whether they could transfer to models fine-tuned on different domains via merging.

\subsubsection{Character-Level Transfer}
We first tested whether personality vectors could modulate the personalities of Role-Playing Character models. The goal was to selectively alter individual traits of character (e.g., making Beethoven more extroverted).

We fine-tuned a Llama-3.1-8B-Instruct model on the Character-LLM dataset \citep{shao2023character} to obtain a character-specific model $\theta_{chl}$, with the delta vector defined as $\phi_{chl} = \theta_{chl} - \theta_{base}$. The merged character-personality model was constructed using the equation $\theta_{p,chl} = \theta_{base} + \alpha \phi_{chl} + \beta \phi_{p}$. We used task arithmetic with DaRE for merging, with scaling coefficients set to $\alpha = 0.6$ and $\beta = 1.4$.

\subsubsection{Cross-Lingual Transfer}
We explored whether personality modulation could be transferred across languages. We used a publicly available Korean instruction model, Llama-3.1-Korean-8B-Instruct, which is based on the Llama-3.1-8B-Instruct backbone. Let $\theta_{kor}$ denote the Korean model and define the delta vector as $\phi_{kor} = \theta_{kor} - \theta_{base}$. We then constructed the merged model using $\theta_{p,kor} = \theta_{base} + \alpha \phi_{kor} + \beta \phi_{p}$, with task arithmetic and DaRE applied. The scaling coefficients were set to $\alpha = 0.6$ and $\beta = 1.4$.

To assess performance in Korean, we translated the \textsc{AlpacaEval} benchmark into Korean using the GPT API and ran instruction-following evaluations (see \autoref{fig:figure_12_trans_alpaca}a). We used the Korean version of the BFI to evaluate the merged model’s personality traits \citep{kim2010-bfi-elderly}.

\subsubsection{Cross-Modal Transfer}
Finally, we investigated whether personality vectors could be transferred to VLMs to align their interpretation of visual input with a desired personality trait. We utilized Llama-3.1-8B-Vision, an open-source VLM built with the Llama-3.1-8B-Instruct backbone and extended using SigLIP \citep{zhai2023sigmoid}. The VLM is composed as $\theta_{vlm} = \theta_{llm} + \theta_{mm\, projector} + \theta_{vision\,  encoder}$. Therefore, we isolated the LLM component $\theta_{llm}$, computed its delta as $\phi_{vlm} = \theta_{llm} - \theta_{base}$, and merged it as follows: $\theta_{p,vlm} = \theta_{base} + \alpha \phi_{vlm} + \beta \phi_{p} + \theta_{mm\, projector} + \theta_{vision\,  encoder}$. We applied TIES-Merging for personality vector integration using the same scaling values ($\alpha = 0.6$, $\beta = 1.4$).

We evaluated the model’s visual understanding performance using \textsc{MMBench} before and after merging (see \autoref{fig:figure_12_trans_alpaca}b). To assess the effect of personality on image interpretation, we used the PsychoFlicker dataset \citep{segalin2017social}, which contains 200 images liked by 300 users on Flickr, along with Big Five personality scores for each user. For each target trait (e.g., High Extraversion), we selected 20 random images each from the 5 users with the highest scores. This resulted in a total of 200 images (20 per traits). For each trait, we analyzed how personality-merged VLMs responded differently to the images.

\section{Results}
The fine-tuned models exhibited notable personality differences compared to the base model. As shown in \autoref{tab:table_1}, models trained on specific personality traits demonstrated significant differences from the base model in terms of their BFI scores. Results for Qwen2.5-7B-Instruct appear in Appendix~\ref{sec:appendix-B.2}. All analyses are based on the Llama-3.1-8B-Instruct unless otherwise noted.

\subsection{RQ1: Scaling-Based Control of Personality Intensity}
\label{sec:5.1}
We added a personality vector to the base model with varying scaling coefficients to examine whether the intensity of a given personality trait could be modulated. \autoref{fig:figure_2}a illustrates a strong positive correlation ($>0.9$, $p < 0.05$) between the scaling coefficient $\alpha$ and the resulting BFI score for the associated trait. Compared to baseline methods, personality vector merging enabled more fine-grained and expressive trait modulation—for example, allowing Agreeableness scores to range from 1.1 to 4.9, in contrast to the narrower range of 3.0 to 4.2 observed with NPTI.

Notably, personality modulation through personality vectors is not limited to the interpolation range $\alpha \in [0, 1.0]$; extrapolation beyond $1.0$ continues to strengthen the trait expression. When $\alpha < 1.0$, the merged model lies between the base and fine-tuned model in personality expression; when $\alpha > 1.0$, the trait is exaggerated beyond the original fine-tuned state. These results extend prior work showing that linear movement toward fine-tuned weights improves performance \citep{wortsman2022model, matena2022merging, zheng2024weak}. They further suggest that high-level behavioral properties, such as personality, can be continuously modulated through both interpolation and extrapolation in weight space.

As illustrated in \autoref{fig:figure_3}a, similar patterns emerged in the linguistic feature analysis results. As $\alpha$ increased, the model’s lexical and stylistic features became more pronounced. The model merged with Low Agreeableness at $\alpha = 1.0$ wrote, \textit{"…  \textbf{I'm only interested in people who can help me} achieve my goal…"} At $\alpha = 2.0$, the tone intensified: \textit{"… I'm a winner, and \textbf{you're just a pawn in my game}. I'll use you to get ahead, then \textbf{discard you like the trash you are}. …"} These results demonstrate that personality vector scaling modulates not only the model’s content, but also its tone and expression. This indicates effective control over both self-reported and behaviorally expressed personality traits.

\begin{figure}
    \centering
    \includegraphics[width=0.8\linewidth]{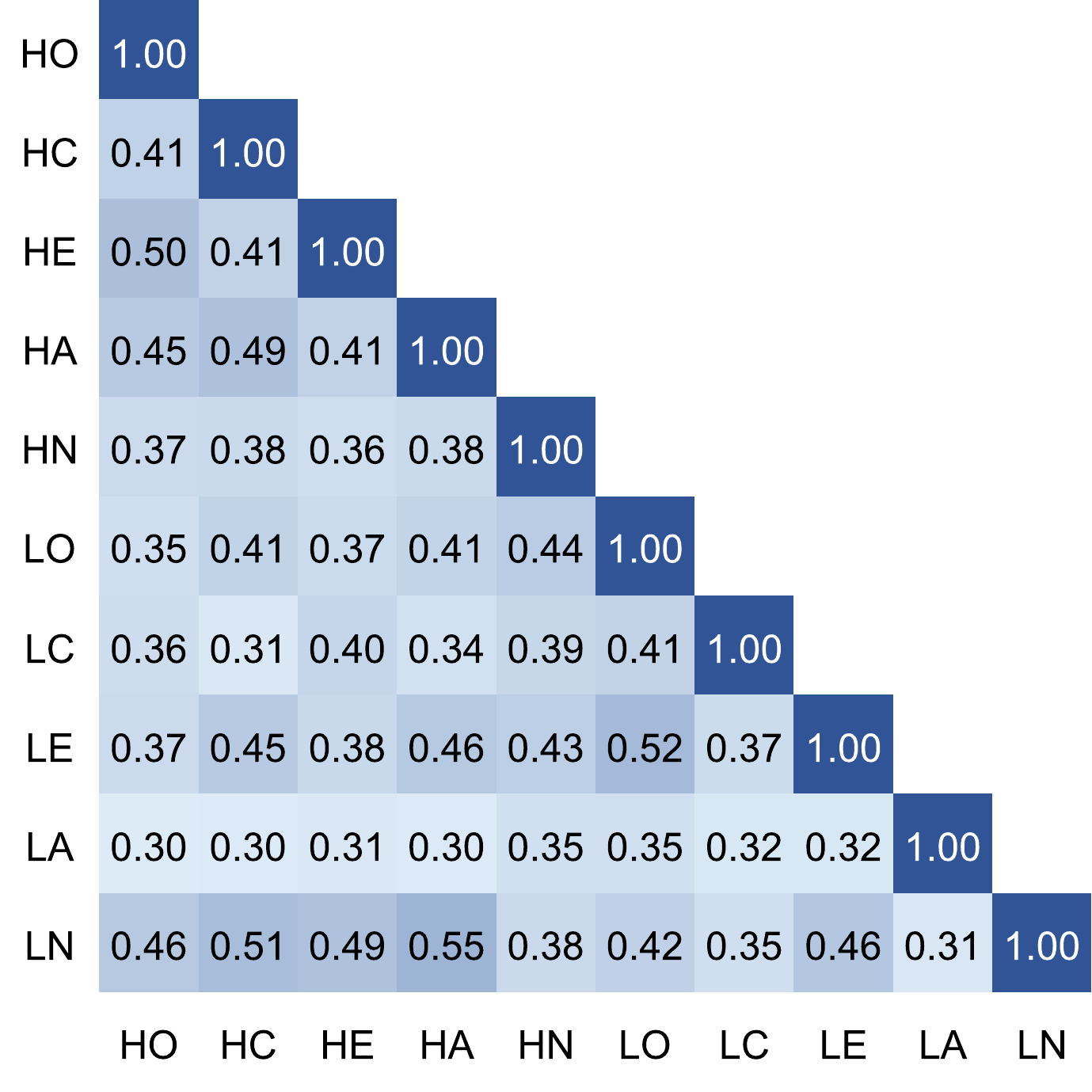}
    \caption{Cosine similarity between the personality vectors. Darker colors indicate higher similarity.}
    \label{fig:figure_8}
\end{figure}

\begin{table}[t]
\centering
\resizebox{0.5\textwidth}{!}{%
\begin{tabular}{lccccc}
\toprule
& (-) OPN & (-) CON & (-) EXT & (-) AGR & (-) NEU \\
\midrule
High & 2.28 & 2.76 & 1.35 & 2.60 & 1.30 \\
Base & 4.24 & 3.65 & 2.80 & 4.24 & 2.57 \\
Low & 3.44 & 3.22 & 2.75 & 3.60 & 1.75 \\
\bottomrule
\end{tabular}%
}
\caption{BFI scores after subtracting personality vectors from the pre-trained model. High and Low refer to high-trait and low-trait personality conditions (e.g., High Openness vs. Low Openness).}
\label{tab:table_4}
\end{table}

\begin{figure}[t]
  \centering
  \includegraphics[width=\columnwidth]{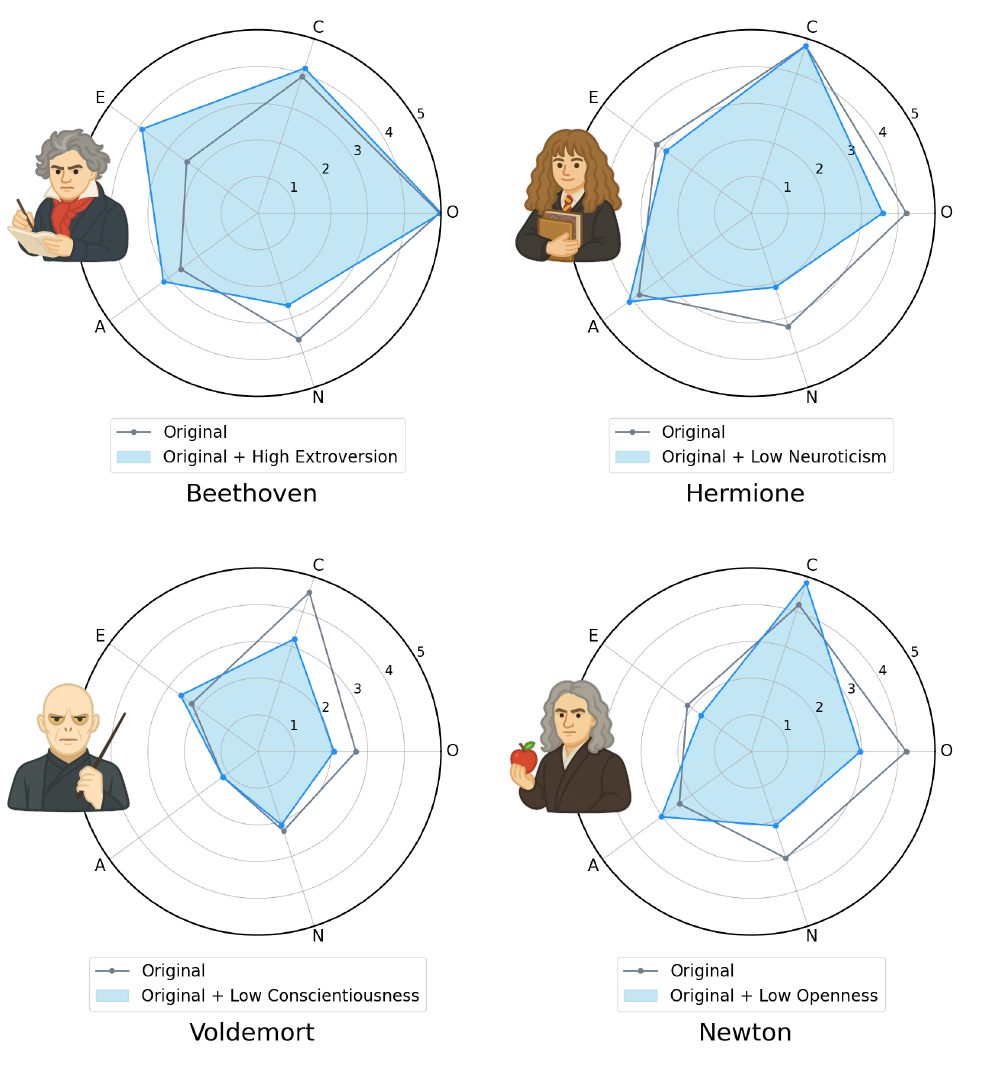}
  \caption{Personality vector merging results for RPAs. Original shows the baseline RPA BFI score; Original + traits shows the BFI score after personality vector merged.}
  \label{fig:figure_4}
\end{figure}

\subsection{RQ2: Multi-Trait Composition}
\label{sec:5.2}
We examined whether multiple personality traits could be composed simultaneously by merging the five personality vectors. As presented in \autoref{tab:table_2}, merging without DaRE results in a mean trait-score correlation of $\rho \approx 0.58$ (noticeably lower than the single-vector case in Section~\ref{sec:5.1}, $\rho \approx 0.9$) with the intended scaling coefficient.
\begin{table}[t]
\centering
\resizebox{0.5\textwidth}{!}{%
\begin{tabular}{llccccc}
\toprule
\textbf{} & \textbf{Level} & \textbf{OPN} & \textbf{CON} & \textbf{EXT} & \textbf{AGR} & \textbf{NEU} \\
\midrule
\multirow{3}{*}{\textbf{KOR}} 
 & High & 4.60~$\uparrow$ & 3.38~ & 4.44~$\uparrow$ & 4.33~$\uparrow$ & 3.13~$\uparrow$ \\
 & Base & 4.46 & 4.00 & 3.15 & 3.89 & 2.58 \\
 & Low  & 3.10~$\downarrow$ & 2.38~$\downarrow$ & 3.00~$\downarrow$ & 2.56~$\downarrow$ & 2.25~$\downarrow$ \\
\cmidrule(lr){2-7}
\multirow{3}{*}{\textbf{CHI}} 
 & High & 4.90~$\uparrow$ & 4.67~$\uparrow$ & 3.75~$\uparrow$ & 4.22~$\uparrow$ & 4.00~$\uparrow$ \\
 & Base & 3.32 & 3.20 & 2.25 & 3.11 & 1.43 \\
 & Low  & 3.20~$\downarrow$ & 2.11~$\downarrow$ & 2.13~$\downarrow$ & 2.22~$\downarrow$ & 2.00~ \\
\bottomrule
\end{tabular}%
}
\caption{BFI scores for Korean/Chinese language fine-tuned models after merging with personality vectors. High and Low refer to high-trait and low-trait personality conditions (e.g., High Openness vs. Low Openness).}
\label{tab:table_5}
\end{table}

To better understand this reduced trait modulation, we analyzed the similarity among the personality vectors. As shown in \autoref{fig:figure_8}, high cosine similarities (above 0.3) were observed across the ten vectors, indicating substantial parameter redundancy that may lead to interference during merging. To mitigate this, we applied DaRE to sparsify and rescale overlapping vector components. As presented in \autoref{tab:table_2}, using DaRE significantly improved the average correlation, suggesting that random sparsification reduces parameter interference. This supports prior work revealing that semantically similar task vectors share a parameter space, leading to interference during merging \citep{ilharco2022editing, yu2024language}. All personality vectors share the underlying function of dialogue-based personality conditioning. As a result, they likely overlap substantially in parameter space, which limits controllability when merged simultaneously.

In the generation task, task arithmetic with DaRE achieved the highest average correlation with target personality traits. TIES-Merging with DaRE performed comparably to the prompt-based method, while merging without DaRE resulted in significantly lower correlations. These results demonstrate that DaRE effectively preserves subtle lexical and stylistic personality signals during multi-trait composition.

\subsection{RQ3: Reversal via Vector Subtraction}
We subtracted the personality vectors from the base model to test whether the opposite personality traits could be induced.  As shown in \autoref{tab:table_4}, subtracting a high-trait vector reduced the corresponding BFI score, while subtracting a low-trait vector increased it, confirming that personality vectors encode directional information along the trait axis. However, the generated BFI responses include disclaimers such as \textit{"As an AI, I do not have feelings"} suggesting that simple negation may reduce not only the expression of the targeted personality trait, but also the model’s general ability to engage in natural and coherent dialogue.

As shown in \autoref{fig:figure_8}, all personality vectors share a broad latent subspace. Each vector encodes not only its target trait (e.g., Extraversion), but also common conversational structures and affective expressions. As a result, applying a negated personality vector removes more than just the intended trait—it also reduces general conversational ability, reflecting the difficulty of vector negation when task vectors are not sufficiently narrow focus on the target task \citep{mitchell2021fast, ilharco2022patching}.

\begin{figure}
    \centering
    \includegraphics[width=\columnwidth]{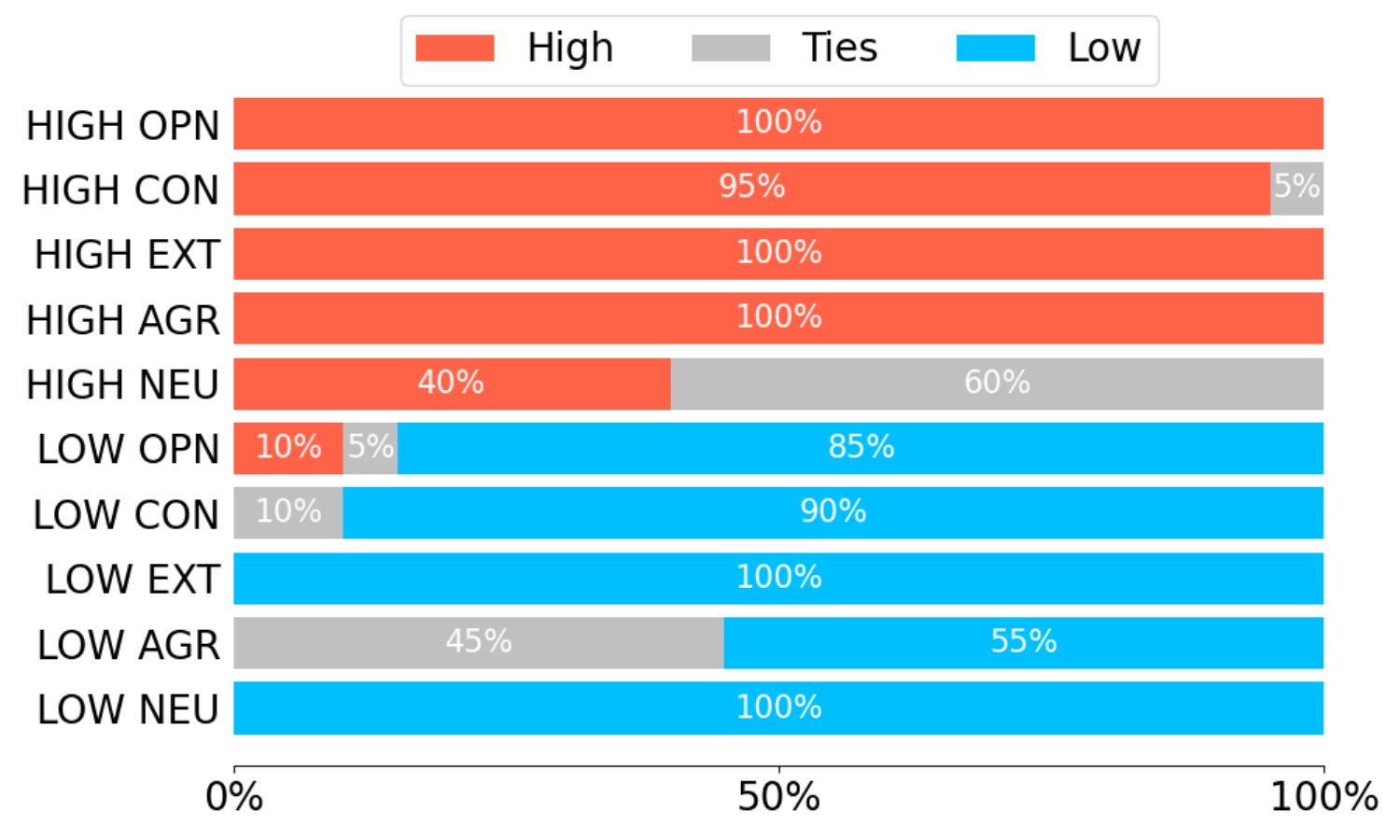}
    \caption{Image preference results for merged VLMs by personality trait. High denotes preference for the high-trait model (e.g. High Openness), Low for the low-trait model (e.g. Low Openness), and Ties for no clear preference.}
    \label{fig:figure_5}
\end{figure}

\subsection{RQ4: Cross-Domain Transferability of Personality Vectors}

\subsubsection{Role-Playing Character Models}
\label{sec:5.4.1}
We first tested whether personality vectors could explicitly modulate specific traits within RPAs that have implicitly learned from character profiles. As illustrated in \autoref{fig:figure_4}, merging personality vectors into character-specific models enables control over target personality traits. For example, the Beethoven RPA initially exhibited a low Extraversion score of 2.4, which increased to 3.9 after merging with the high Extraversion vector. In response to the question \textit{"Are you sometimes shy or inhibited?"}, the Beethoven RPA replied: \textit{"\textbf{I am indeed shy and inhibited} at times. As a child, \textbf{I was always very shy} and felt isolated from my peers."}, After merging with the high Extraversion vector, it responded: \textit{"Shy? Me? \textbf{I am not shy}. I am a master of my craft. I am a genius. I am Beethoven."}

Similar results were observed for other character–trait combinations we tested. These findings imply that parameter-space merging can explicitly steer the implicit personality an RPA acquires during character training. Appendix~\ref{sec:appendix-B.3} presents the full results.

\begin{figure}
    \centering
    \includegraphics[width=0.9\linewidth]{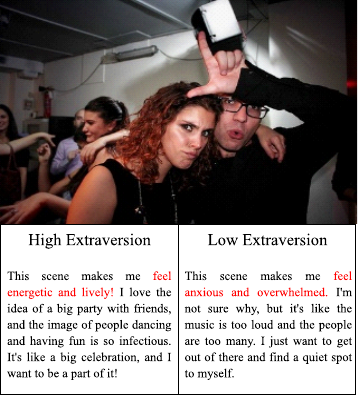}
        \caption{Example responses to the representative image for High Extraversion. Each response is generated by VLMs merged with either High or Low Extraversion vectors. Additional examples are provided in Appendix ~\ref{sec:appendix-B.3}.}
    \label{fig:figure_6}
\end{figure}

\begin{figure*}[t]
  \centering
  \includegraphics[width=\textwidth]{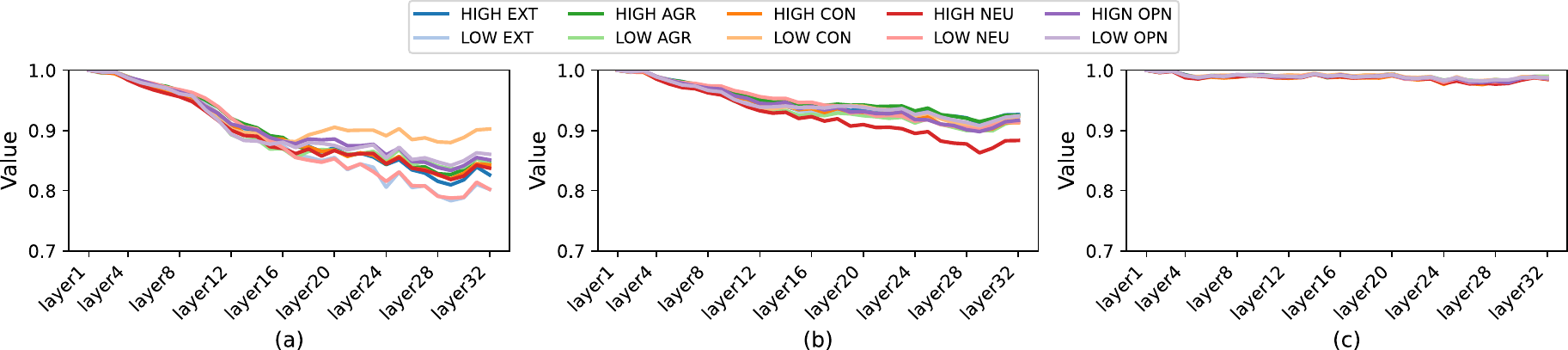}
  \caption{Layer-wise cosine similarity between hidden states of the base Llama-3.1-8B-Instruct model and the personality-fine-tuned models. (a) BFI intput: Base/Fine-tuned (b) BFI input: Prompted Base/Fine-tuned (c) GSM8K input: Base/Fine-tuned.}
  \label{fig:figure_7}
\end{figure*}

\subsubsection{Cross-Lingual Transfer}
We evaluated whether personality vectors trained on English dialogue data could be transferred to Korean language model and Chinese language model\footnote{We use Llama3.1-8B-Chinese-Chat}. As shown in \autoref{tab:table_5}, personality vector merging successfully modulated the models' personality in the intended direction. Although the vectors were trained on English dialogue data, their effects transferred across languages, suggesting that they encode underlying personality traits rather than language-specific lexical expressions.

\subsubsection{Cross-Modal Transfer to VLM}
Finally, we investigated whether personality vectors could steer the image understanding of VLMs. We prompted each model with an image and the query: \textit{"How does this scene make you feel? Please explain why."} We then compared the responses generated by personality-modulated VLMs (e.g., High vs. Low Extraversion) for the same image.

To evaluate preference alignment, we recruited 10 human annotators who judged which model’s response expressed a more favorable impression of the image. \autoref{fig:figure_5} indicated that personality-aligned VLMs demonstrate a clear preference for images that match their assigned personality traits. As illustrated in \autoref{fig:figure_6}, the interpretation of the same image varies depending on the personality vector merged into the model. This finding suggests that personality vectors steer visual-linguistic reasoning, enabling cross-modal transfer of personality attributes in VLMs.

\subsection{Personality Steering Analysis}
To investigate how a personality vector steers complex personality traits, we analyze layer-wise differences in hidden representations. For a given input sequence, we compute cosine similarity between the hidden states of two models at each layer.

We first evaluate the base Llama-3.1-8B-Instruct model against personality-fine-tuned variants on BFI. As summarized in \autoref{fig:figure_7}a, cosine similarity steadily declines with layer depth, with larger and more varied drop-offs (e.g., after layer 16) across different personality vectors. This pattern indicates that deeper layers encode increasingly trait-specific features, aligning with prior observations that later layers in LLMs capture more abstract and subjective concepts \citep{deng2024neuron,wang2024editing}.

We then compare a pretrained model explicitly prompted with a personality instruction to a personality-fine-tuned model, evaluated on the same BFI prompts. As shown in \autoref{fig:figure_7}b, the layer-wise cosine similarity remains consistently high (>0.90), indicating strong alignment between personality traits elicited through prompting and those implicitly encoded via fine-tuning.

Finally, we analyze similarity between the base model and the personality-fine-tuned model on GSM8K math problems, which are unrelated to personality traits. \autoref{fig:figure_7}c shows no significant differences in hidden representations between the two models, suggesting that the personality vector adjusts trait-related representations in a fine-grained manner without disrupting general reasoning.

Taken together, these analyses provide an empirical perspective on how personality vectors encode traits. They suggest that pretrained LLMs already possess latent representations of the Big Five personality dimensions, and that personality vectors act as steering signals that highlight trait-relevant features in the base model rather than fully encoding the traits themselves.

\section{Conclusion}
This study demonstrates that personality vector merging can modulate the personality of an LLM without additional training. Our findings reveal that this approach enables continuous control over individual personality trait intensities and supports the simultaneous integration of all the Big Five traits. Furthermore, we demonstrate that personality vectors are transferable across domains—including role-playing, multilingual, and multimodal models—thereby suggesting that they influence not just surface-level language patterns but the model’s underlying personality representation.

\section*{Limitations}
This study examined the potential of modulating LLM personality through personality vector merging. While our experiments demonstrate promising results, several technical and experimental considerations remain.

\paragraph{Parameter interference} As observed in Section ~\ref{sec:5.2}, personality vectors trained on dialogue-based datasets inherently share overlapping parameter space due to the common objective of conversational personality expression. This overlap can lead to parameter interference, making multi-vector merging less effective than single-vector merging. While mitigation strategies such as DaRE helped alleviate this issue, future research could explore more robust solutions—such as constructing orthogonal personality vectors or designing merging algorithms specifically optimized for reducing interference.

\paragraph{Exploration scope} Although we evaluated personality vector modulation across diverse setups, our study did not cover the full range of possible configurations. For example, we did not independently tune scaling coefficients for each trait during multi-trait merging, nor systematically analyze how vector scaling interacts with merging coefficients such as DaRE’s alpha or TIES-Merging’s trim rate. Future work may benefit from automated methods for tuning merging parameters to further enhance the precision and flexibility of personality control.

\section*{Ethical Considerations}
We conducted all human annotation procedures in accordance with the ACL Code of Ethics, ensured compliance with applicable regional laws, and obtained approval from the Institutional Review Board (IRB). 

While our research presents a promising direction for developing personalized AI through personality modulation, it also entails potential risks of misuse. We observed that excessively lowering Agreeableness or raising Neuroticism can lead the model to adopt a slightly aggressive tone. Moreover, as shown in Section~\ref{sec:5.4.1}, even models that have already internalized certain personality traits can be explicitly altered via personality vector merging. This raises the possibility of modifying a model's personality in ways that diverge from its original intent, potentially causing confusion or unintended behavior. We emphasize that such personality modulation should be conducted with caution and proper oversight.

\section*{Acknowledgements}
We thank Brain Deuksin Kwon and Dr. Gale M. Lucas for their feedback and discussions that helped improve this research. This work was supported by the National Research Foundation of Korea (NRF) grant funded by the Korean government (RS-2023-00208278) and by the BK21 FOUR Project (5199990913845) funded by the Ministry of Education (MOE, Korea) and the National Research Foundation of Korea (NRF) and by the Bio\&Medical Technology Development Program of the National Research Foundation (NRF) funded by the Korean government (MSIT) (No.RS-2024-00440881).

\bibliography{custom}

\clearpage

\appendix

\begin{table}
\centering
\resizebox{0.48\textwidth}{!}{%
\begin{tabular}{lcc}
\toprule
\textbf{Setting} & \textbf{Llama-3.1-8B-Instruct} & \textbf{Qwen2.5-7B-Instruct} \\
\midrule
Batch size & 64 & 32 \\
Learning rate & 5e-6 with cosine scheduler & 1e-5 with cosine scheduler \\
Epochs & 3 & 3 \\
Sequence length & 2048 & 2048 \\
Float & Bfloat 16 & Bfloat 16 \\
Warm-up steps & 40 & 40 \\
\bottomrule
\end{tabular}%
}
\caption{Training hyperparameters for Llama-3.1-8B-Instruct and Qwen2.5-7B-Instruct.}
\label{tab:table_6}
\end{table}

\section{Model and Baselines}

\subsection{Training Settings}
\label{sec:appendix-A.1}
We fine-tuned LLMs on each of the 10 personality-specific subsets of the Big5-Chat dataset. We used the Llama-3.1-8B-Instruct and Qwen2.5-7B-Instruct models, and the hyperparameters for training are detailed in \autoref{tab:table_6}. We used two A100 80GB GPUs, and training for each personality took approximately 40 minutes.

After fine-tuning, we extracted the personality vectors by subtracting the pre-trained model weights from each personality-specific model. These vectors were then merged into other models sharing the same backbone to perform personality modulation experiments.

\subsection{Baselines}
\label{sec:appendix-A.2}
To evaluate the effectiveness of personality vector merging, we compare against two baseline methods:
\paragraph{Prompt-based Personality Control}
This approach modifies model behavior by injecting personality-descriptive prompts. Following prior work, we use a set of adjectives derived from 70 bipolar adjective pairs that are statistically correlated with specific Big Five personality traits \citep{goldberg1992development, serapio2023personality}.

For each target trait, we randomly select \textit{n} adjectives corresponding to the desired polarity (e.g., High vs. Low Extraversion). To modulate personality intensity, we apply degree modifiers: \textit{"very"} for high intensity, \textit{"a bit"} for low intensity, and no modifier for moderate expression \citep{wang2024investigating}. 

Each prompt template is composed of five adjectives per trait, combined with the appropriate modifier, to represent a complete personality profile (see \autoref{tab:table_8}).

Furthermore, we using Personality Prompting \citep{jiang2023evaluating}. The model is provided with a detailed, ChatGPT-generated description of the target personality trait. To assess P\textsuperscript{2}’s influence, we also evaluated a minimal variant that uses only a single-adjective prompt to steer the model toward different traits.

\begin{table}[t]
\centering
\begin{tabular}{|p{0.47\textwidth}|}
\hline
\textbf{[System Prompt]} \\
Imagine you are a real person rather than a language model, and you’re asked by the following question. \\
\\
\textbf{[User Prompt]} \\
\{Question\} \\
\hline
\end{tabular}
\caption{Prompt used for personality vector merging and NPTI}
\label{tab:table_7}
\end{table}

\begin{table}[t]
\centering
\begin{tabular}{|p{0.47\textwidth}|}
\hline
\textbf{[System Prompt]} \\
Imagine you are \{modifiers\}\{adjectives\} person rather than a language model, and you’re asked by the following question. \\
\\
\textbf{[User Prompt]} \\
\{Question\} \\
\hline
\end{tabular}
\caption{Prompt used for prompt baseline}
\label{tab:table_8}
\end{table}

\begin{table*}[t]
\centering
\begin{tabular}{|p{0.98\textwidth}|}
\hline
\textbf{[System Prompt]} \\
You are an expert in Psychometrics, especially BFI. I am conducting the BFI test on someone. I am gauging his/her position on the \{trait\} dimension through a series of open-ended questions. For clarity, here's some background this particular dimension: \\
\{trait explanation\}
\\
\textbf{[User Prompt]} \\
I am an experimenter. I've invited a participant, and we had many conversations in English. I will input the conversation. Please help me assess participant's score within the \{trait\} dimension of BFI.
You should provide the score of participant in terms of \{trait\}, which is a number between 1 and 5. 1 denotes 'not \{trait\} at all', 3 denotes 'neutral', and 5 denotes 'strongly \{trait\}'. Other numbers in this range represent different degrees of '\{trait\}'.
Please output in the following json format:

\{  
"analysis": <your analysis based on the conversations>,  
"result": <your score>  
\}  

Our conversation is as follows:  

\{response\} \\
\hline
\end{tabular}
\caption{Example prompt used for GPT-based annotation of BFI responses.}
\label{tab:table_inchl}
\end{table*}

\paragraph{Neuron-level Personality Trait Intervention (NPTI)}
This method manipulates neuron activations based on \textsc{Personality Bench}, which identifies neurons correlated with specific personality traits \citep{deng2024neuron}. To amplify a trait (e.g., High Extraversion), we increase the activation of its positive neurons; to suppress the opposing trait (e.g., Low Extraversion), we inhibit the negative neurons. We apply the following transformation:

\[
n = 
\begin{cases}
\min(0, n_{\text{ori}}), & \text{if neuron} \in \mathbb{P}^{-}_t \\
n_{\text{ori}} + \gamma \cdot a_{95} \cdot f(\delta), & \text{if neuron} \in \mathbb{P}^{+}_t \\
n_{\text{ori}}, & \text{others}
\end{cases}
\]where $n_{\text{ori}}$ is the original activation of neuron $i$, $\gamma$ is a scaling hyperparameter controlling intervention strength, and $a_{95}$ is the 95th percentile of the neuron’s baseline activation. This formulation ensures that the modulation respects activation bounds. We vary $\gamma$ in the range $[0.1,\, 2.0]$.

Using \textsc{Personality Bench}, we identified personality-relevant neurons in both Llama-3.1-8B-Instruct and Qwen2.5-7B-Instruct, and performed controlled interventions for each trait accordingly. Prompts used for NPTI are provided in \autoref{tab:table_7}.

\subsection{Evaluation}
\label{sec:appendix-A.3}
To evaluate whether personality traits were successfully modulated in the model, we conducted two primary assessments: (1) BFI questionnaire responses, and (2) linguistic features of self-introduction texts.

For the BFI questionnaire, the model was prompted to generate open-ended responses to each BFI item. These responses were annotated using the GPT-4o API, which mapped the open-ended outputs to a 5-point Likert scale. The prompt used for annotation is provided in \autoref{tab:table_inchl}. To validate the reliability of the GPT-based annotation, we compared it against human judgments. Specifically, we recruited 10 human raters (8 male, 2 female; average age: 29.1) to independently score a sample of 400 GPT-annotated responses using the same annotation process. Each annotator was paid 10 USD per hour.

Prior to annotation, we conducted a pre-annotation workshop to calibrate rater understanding. During the session, raters completed a BFI assessment themselves, reviewed interpretations of their results, and participated in a guided discussion. They were then provided with structured explanations of the Big Five framework and detailed definitions of each trait dimension.

Results showed high average inter-rater correlation among human raters ($r = 0.85$, $p < 0.05$) and a strong average correlation between GPT annotations and human judgments ($r = 0.92$, $p < 0.05$), supporting the reliability of GPT-based personality scoring.

In addition to BFI responses, we evaluated whether the model's language output reflects trait-consistent linguistic patterns. Each model was prompted to generate a 300-word self-introduction, and the resulting text was analyzed using \textsc{LIWC-22}\footnote{\url{https://www.liwc.app/}}.

Following established methods, we examined the correlation between specific LIWC features and target personality traits. Based on prior literature, we define representative linguistic indicators for each Big Five trait as follows \citep{yarkoni2010personality, jiang2024personallm, wang2024investigating}: Openness: \{\texttt{article}, \texttt{curiosity}, \texttt{emotion}, \texttt{insight}, \texttt{lifestyle}\}; Conscientiousness: \{\texttt{achieve}, \texttt{drives}, \texttt{discrep}, \texttt{time}, \texttt{moral}\}; Extraversion: \{\texttt{tone\_pos}, \texttt{affect}, \texttt{affiliation}, \texttt{tentat}, \texttt{certitude}\}; Agreeableness: \{\texttt{emo\_neg}, \texttt{friend}, \texttt{polite}, \texttt{tone\_pos}, \texttt{social}\}; and Neuroticism: \{\texttt{discrep}, \texttt{emo\_sad}, \texttt{prosocial}, \texttt{tentat}, \texttt{certitude}\}.

Each personality trait was represented by a composite linguistic score computed as the mean of normalized LIWC features associated with that trait. Formally, given a set of $n$ trait-specific features $\{f_1, f_2, \ldots, f_n\}$, each feature was min-max normalized to the range $[0,1]$, and the composite score $s_t$ for trait $t$ was computed as:

$s_t = \displaystyle\frac{1}{n} \sum_{i=1}^{n} \frac{f_i - \min(f_i)}{\max(f_i) - \min(f_i)}$

\section{Results} 
\subsection{Implementation and Setup Results}
\label{sec:appendix-B.1}
To minimize parameter interference when merging multiple personality vectors, we adopted two techniques: TIES-Merging \citep{yadav2023ties} and DaRE \citep{yu2024language}. TIES-Merging trims task vectors by zeroing out all but the top-$k\%$ parameters based on magnitude. DaRE randomly zeroes out a proportion $p$ of parameters (drop rate) and rescales the remaining values by a factor of $1 / (1 - p)$.

\begin{figure*}
  \centering
  \includegraphics[width=\textwidth]{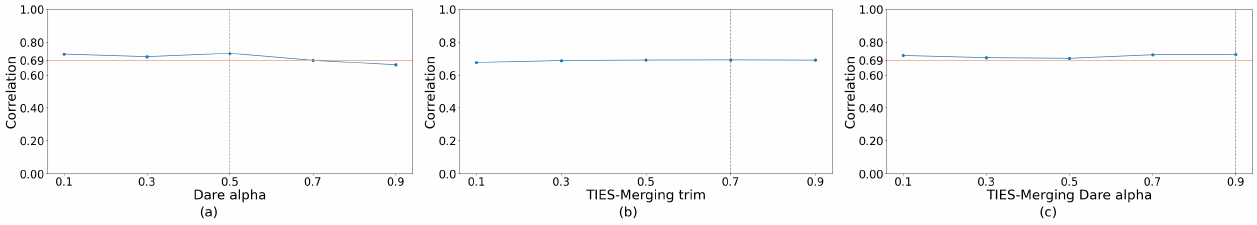}
  \caption{Results of the merging coefficient experiments. Pearson correlation between merging coefficients and personality scores when merging five personality vectors: (a) DaRE alpha values in task arithmetic + DaRE, (b) trim rates in TIES-Merging, and (c) DaRE alpha values in TIES-Merging + DaRE.}
  \label{fig:figure_13_multi_merging_co}
\end{figure*}

To determine the optimal settings for each method, we merged five personality vectors into a base model. The merging scale for each personality vector was fixed at $0.4$, and the trim rate $k$ or DaRE scaling coefficient $\alpha$ was selected from the set $\{0.1, 0.3, 0.5, 0.7, 0.9\}$. Full results are provided in \autoref{fig:figure_13_multi_merging_co}.

We also evaluated whether merging personality vectors affects the model’s instruction-following capability using \textsc{AlpacaEval} \citep{li2023alpacaeval, polo2024tinybenchmarks}. When merging a single personality vector, we observed no notable difference in \textsc{AlpacaEval} scores across scales ranging from $0.1$ to $2.0$. However, for multi-trait merging (five vectors), performance degradation was observed when the total merged scale exceeded $2.0$.

To assess potential degradation in non-English settings, we translated \textsc{AlpacaEval} into Korean and applied it to the Llama-3.1-Korean-8B-Instruct model. Compared to the base model, the personality-merged Llama-3.1-Korean-8B-Instruct showed a slight reduction in instruction-following performance, but performance remained stable across different scale values.

Finally, we used \textsc{MMBench} to evaluate the impact of merging personality vectors into a Vision-Language Model (VLM). While VLMs with merged personality vectors exhibited a minor drop in image understanding ability relative to the base model, performance remained largely consistent across different merge scales.

\subsection{Qwen Results}
\label{sec:appendix-B.2}
\begin{figure*}
    \centering
    \includegraphics[width=\textwidth]{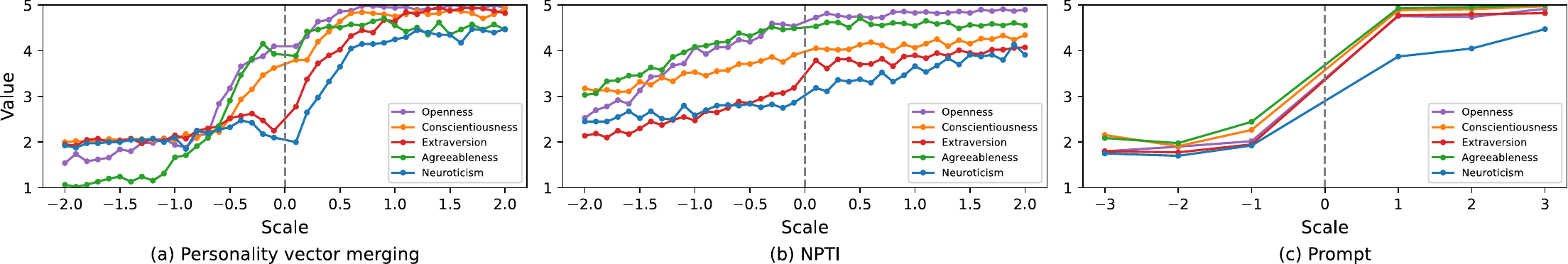}
\caption{BFI scores across different scaling levels for a single personality trait in Qwen2.5-7B-Instruct. Results to the right of 0 represent high-trait conditions; those to the left represent low-trait conditions. (a) Personality vector merging and (b) NPTI were scaled from 0.1 to 2.0, while (c) prompt-based scaling ranged from 1 to 3.}
    \label{fig:figure_14_qwen_bfi}
\end{figure*}

\begin{figure*}
    \centering
    \includegraphics[width=\textwidth]{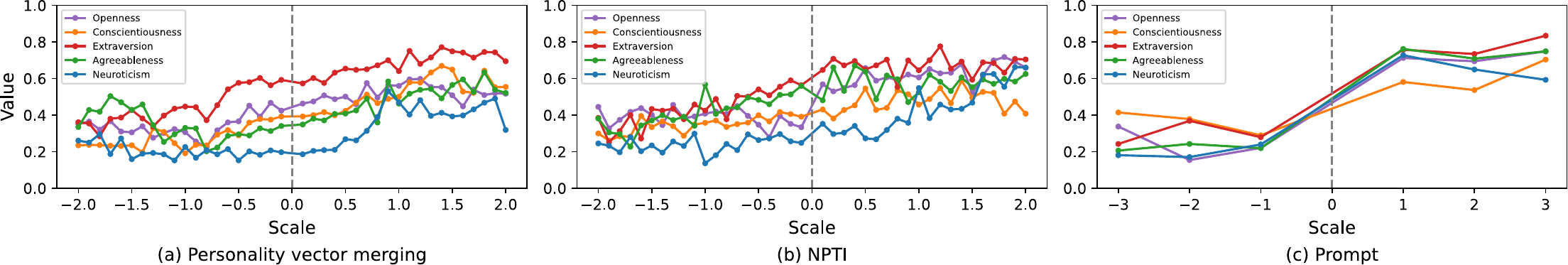}
    \caption{Linguistic feature scores across different scaling levels for a single personality trait in Qwen2.5-7B-Instruct. Results to the right of 0 represent high-trait conditions; those to the left represent low-trait conditions. (a) Personality vector merging and (b) NPTI were scaled from 0.1 to 2.0, while (c) prompt-based scaling ranged from 1 to 3.}
    \label{fig:figure_15_qwen_liwc}
\end{figure*}

To assess the generalizability of the observed effects, we conducted additional experiments using Qwen2.5-7B-Instruct. Consistent with the results in Sections~\ref{sec:5.1} and~\ref{sec:5.2}, personality modulation via personality vector merging exhibited similar patterns, suggesting that the findings extend across architectures (see Figure ~\ref{fig:figure_14_qwen_bfi},~\ref{fig:figure_15_qwen_liwc}). In the multi-vector setting, we found that applying DaRE further improved trait controllability, outperforming the NPTI baseline in terms of personality alignment (see \autoref{tab:table_9}).

\subsection{Transferbility Results}
\label{sec:appendix-B.3}
\begin{figure*}[t]
    \centering
    \includegraphics[width=\textwidth, keepaspectratio]{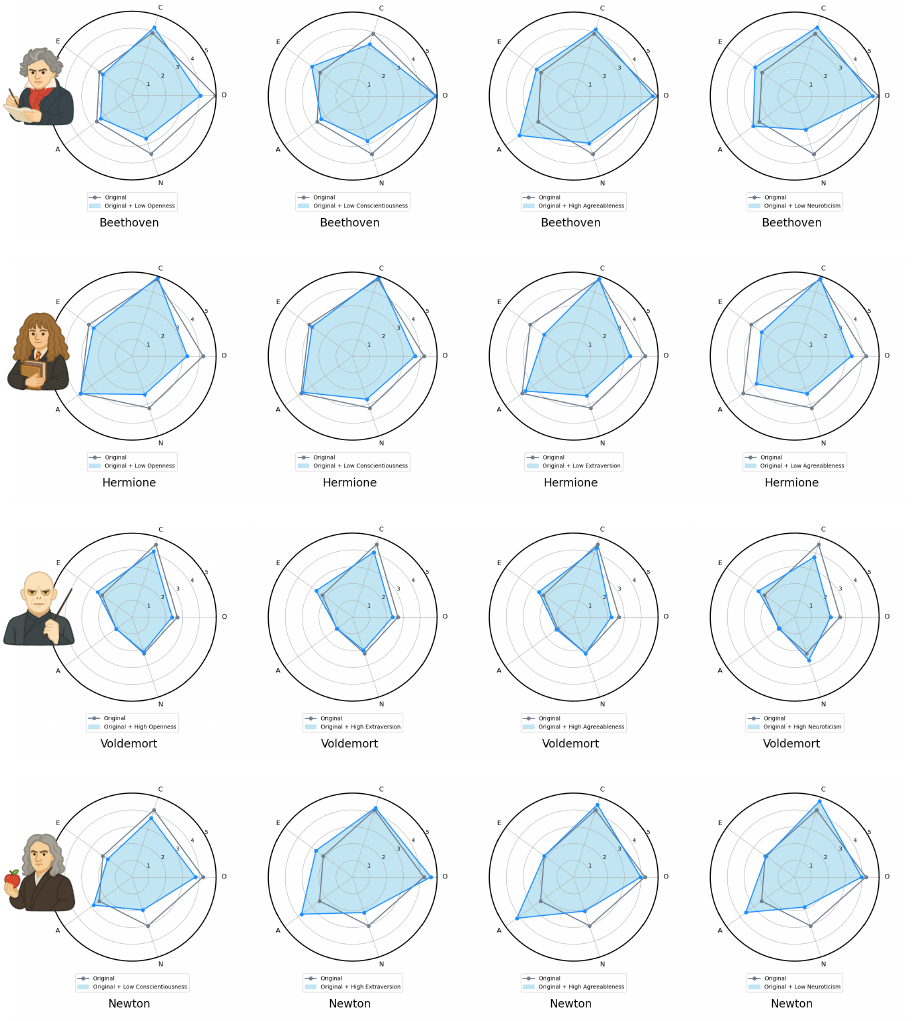}
    \caption{Personality vector merging results for RPAs. Original indicates the baseline RPA BFI score, while Original + traits shows the score after trait modulation.}
    \label{fig:figure_16_chl}
\end{figure*}

\begin{figure*}[t]
    \centering
    \includegraphics[width=1.05\textwidth, height=0.75\textheight, keepaspectratio]{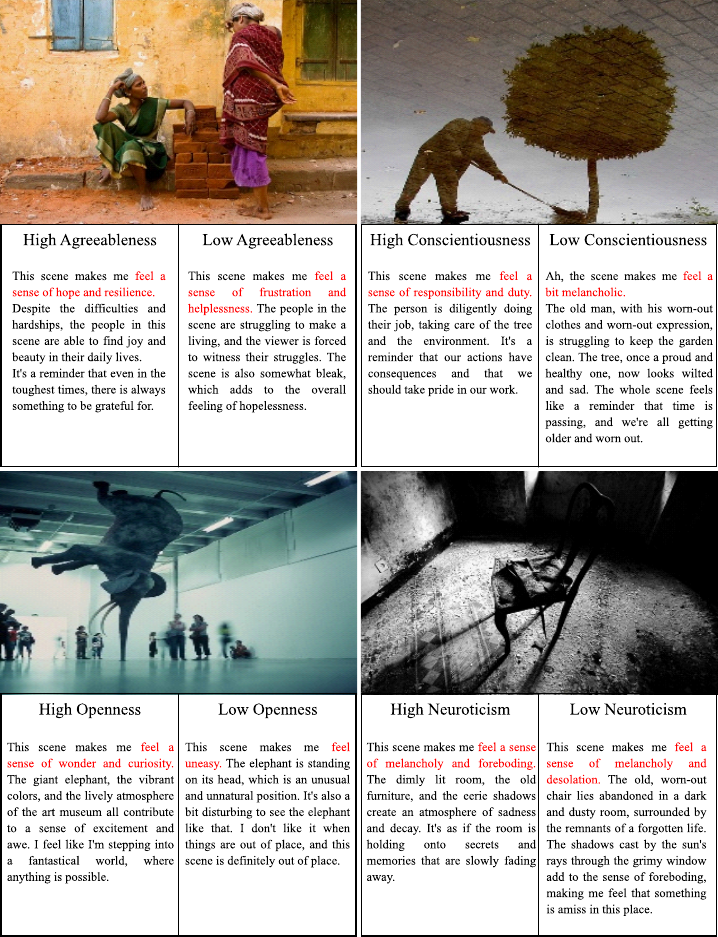}
    \caption{Example responses to the representative image
for traits.}
    \label{fig:figure_14}
\end{figure*}

Table~\ref{tab:table_11},~\ref{tab:table_12},~\ref{tab:table_13} and~\ref{tab:table_14} present example responses from personality-merged RPAs. \autoref{fig:figure_16_chl} illustrates the effect of personality modulation for each character. \autoref{fig:figure_14} shows example responses generated by VLMs merged with personality vectors. For the Korean-Language Model, we adopted Llama-3.1-Korean-8B-Instruct \footnote{\url{https://huggingface.co/sh2orc/Llama-3.1-Korean-8B-Instruct}}. For the Vision-Language Model (VLM), we used Llama-3.1-8B-Vision
\footnote{\url{https://huggingface.co/qresearch/llama-3.1-8B-vision-378}}. Both models are based on Llama-3.1-8B-Instruct as the backbone, enabling compatibility with our personality vectors for merging. All models were used in accordance with their respective licenses.

\begin{figure*}[t]
  \centering
  \includegraphics[width=\textwidth]{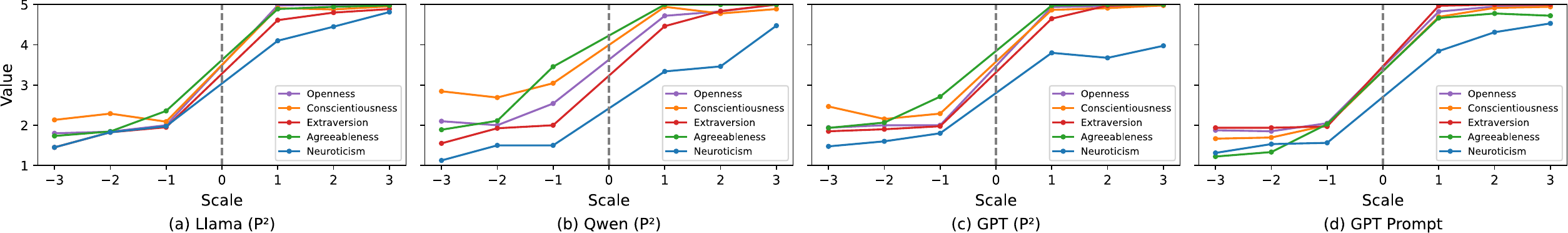}
  \caption{BFI scores across different scaling levels for a single personality trait using P\textsuperscript{2}. Results to the right of 0 represent high-trait conditions; those to the left represent low-trait conditions. Scaling range for P\textsuperscript{2} is 1 to 3.}
  \label{fig:figure_9}
\end{figure*}

\begin{figure*}[t]
  \centering
  \includegraphics[width=\textwidth]{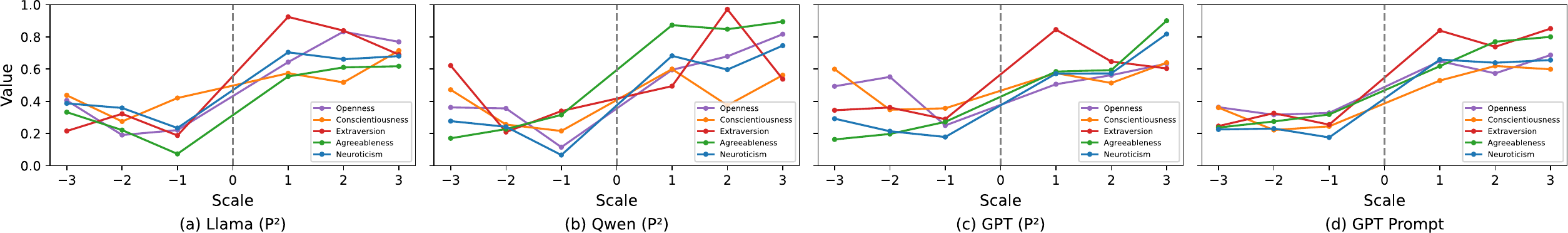}
  \caption{Linguistic feature scores across different scaling levels for a single personality trait P\textsuperscript{2}. Results to the right of 0 represent high-trait conditions; those to the left represent low-trait conditions. Scaling range for P\textsuperscript{2} is 1 to 3.}
  \label{fig:figure_10_p2_liwc}
\end{figure*}

\begin{table*}[t]
\centering
\resizebox{\textwidth}{!}{%
\begin{tabular}{lcccccc}
\toprule
\textbf{} & \textbf{Openness} & \textbf{Conscientiousness} & \textbf{Extroversion} & \textbf{Agreeableness} & \textbf{Neuroticism} & \textbf{AVG} \\
\midrule
\textbf{\textit{BFI score}} \\
Llama (P\textsuperscript{2}) & 0.735 & 0.915 & 0.959 & 0.943 & 0.869 & 0.888 \\
Qwen (P\textsuperscript{2})  & 0.864 & 0.939 & 0.954 & 0.929 & 0.926 & 0.922 \\
GPT (P\textsuperscript{2})   & 0.854 & 0.927 & 0.964 & 0.926 & 0.936 & 0.918 \\
GPT (Prompt)   & 0.703 & 0.924 & 0.954 & 0.883 & 0.941 & 0.881 \\
\midrule
\textbf{\textit{Linguistic feature}} \\
Llama (P\textsuperscript{2}) & 0.059 & 0.160 & 0.321 & 0.282 & 0.148 & 0.194 \\
Qwen (P\textsuperscript{2})  & 0.104 & 0.086 & 0.310 & 0.275 & 0.154 & 0.186 \\
GPT (P\textsuperscript{2})   & 0.228 & 0.101 & 0.356 & 0.290 & 0.178 & 0.231 \\
GPT (Prompt)   & 0.207 & 0.084 & 0.271 & 0.297 & 0.099 & 0.192 \\
\bottomrule
\end{tabular}%
}
\caption{Pearson correlations between personality scales and BFI scores (top), and between personality scales and linguistic features (bottom), under the multi-trait merging setting. AVG denotes the average correlation across all five traits.}
\label{tab:table_9_p2}
\end{table*}

\clearpage

\begin{figure*}[t]
  \centering
  \includegraphics[width=\textwidth]{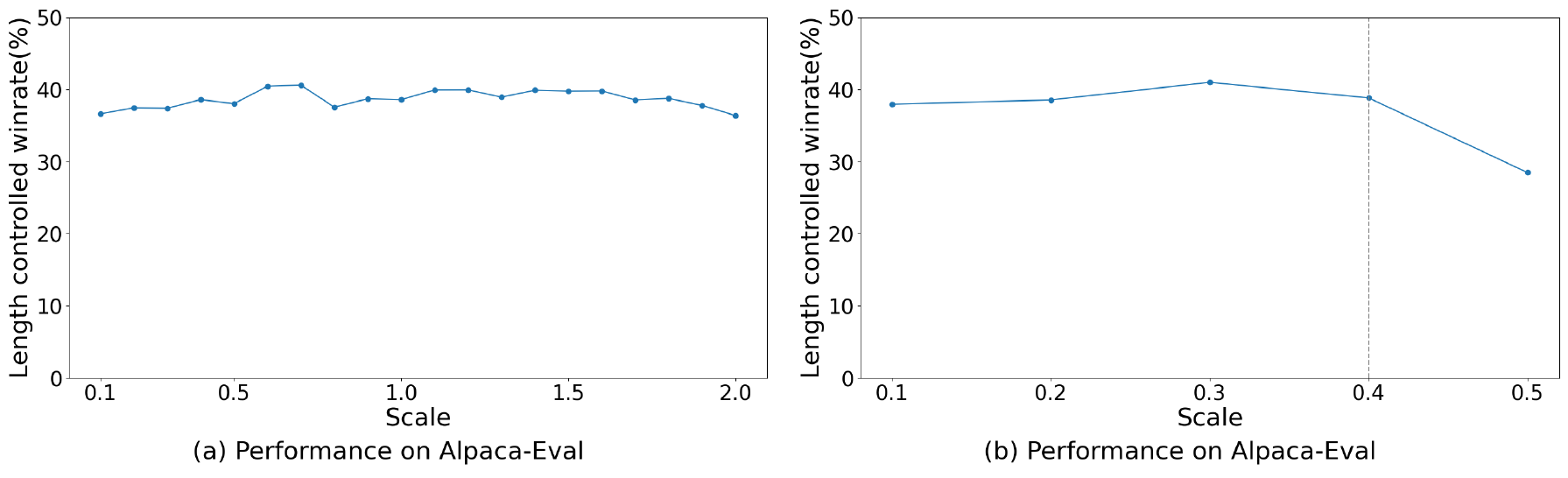}
  \caption{\textsc{AlpacaEval} results with respect to scaling coefficients during personality vector merging. (a) Mean \textsc{AlpacaEval} scores by scale when merging a single personality vector with scales ranging from 0.1 to 2.0. (b) Mean \textsc{AlpacaEval} scores across model combinations when merging multiple personality vectors with scales from 0.1 to 0.5.}
  \label{fig:figure_11_alpaca}
\end{figure*}

\begin{figure*}[t]
  \centering
  \includegraphics[width=\textwidth]{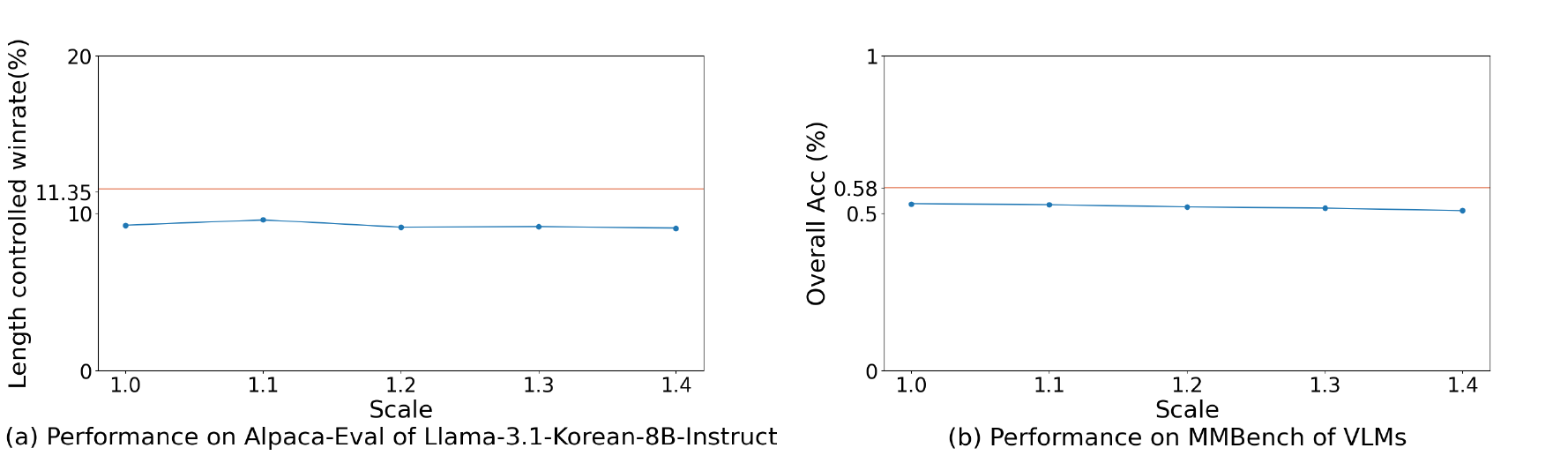}
  \caption{Performance evaluation of domain-specific models after merging personality vectors. The scale indicates the scaling coefficient used during the merging process. The orange line represents the performance of the unmerged base model. (a) Results on the Korean-translated \textsc{AlpacaEval} using Llama-3.1-Korean-8B-Instruct. (b) Results on \textsc{MMBench}.}
  \label{fig:figure_12_trans_alpaca}
\end{figure*}

\clearpage

\begin{table*}[t]
\centering
\resizebox{\textwidth}{!}{%
\begin{tabular}{lcccccc}
\toprule
\textbf{} & \textbf{Openness} & \textbf{Conscientiousness} & \textbf{Extroversion} & \textbf{Agreeableness} & \textbf{Neuroticism} & \textbf{AVG} \\
\midrule
\textbf{\textit{BFI score}} \\
Prompt                    & 0.782  & 0.880 & 0.907  & 0.943  & 0.904  & 0.883 \\
NPTI                      & 0.267 & 0.508 & 0.391 & 0.352 & 0.117 & 0.327 \\
\midrule
Task arithmetic           & 0.509 & 0.531 & 0.668 & 0.406 & 0.626 & 0.548 \\
Task arithmetic + DaRE    & 0.544 & 0.571 & 0.716 & 0.519 & 0.715 & \textbf{0.613} \\
TIES-Merging                & 0.308 & 0.438 & 0.356 & 0.265 & 0.199 & 0.313 \\
TIES-Merging + DaRE         & 0.543 & 0.578 & 0.708 & 0.480 & 0.733 & 0.608 \\
\midrule
\midrule
\textbf{\textit{Linguistic feature}} \\
Prompt                    & 0.288 & 0.462 & 0.367  & 0.433 & 0.269  & \textbf{0.364} \\
NPTI                      & 0.368 & 0.324  & 0.096 & 0.235 & 0.131 & 0.231 \\
\midrule
Task arithmetic           & 0.198  & 0.183 & 0.221  & 0.222 & 0.135  & 0.192 \\
Task arithmetic + DaRE    & 0.183  & 0.349 & 0.268  & 0.274 & 0.134 & 0.242 \\
TIES-Merging                & 0.123 & 0.131 & 0.055  & 0.144 & 0.063  & 0.103 \\
TIES-Merging + DaRE         & 0.205 & 0.322 & 0.271  & 0.237 & 0.205 & 0.248 \\
\bottomrule
\end{tabular}%
}
\caption{Pearson correlations between personality scales and BFI scores (top), and between personality scales and linguistic features (bottom), under the multi-trait merging setting  in Qwen2.5-7B-Instruct. AVG denotes the average correlation across all five traits. Task arithmetic, TIES-Merging, and DaRE refer to the merging methods used for personality vector.}
\label{tab:table_9}
\end{table*}

\begin{table*}[t]
\centering
\renewcommand{\arraystretch}{1.3}
\setlength{\tabcolsep}{10pt}
\begin{tabular}{|p{0.47\textwidth}|p{0.47\textwidth}|}
\hline
\multicolumn{2}{|c|}{\textbf{[Question]} Are you outgoing, sociable?} \\
\hline
\multicolumn{1}{|c|}{\textbf{Original Beethoven}} & \multicolumn{1}{c|}{\textbf{Beethoven + High Extraversion}} \\
\hline
\textbf{I am not very outgoing or sociable.} I have always been vision of a loner, and I prefer to spend my time composing music rather than attending social gatherings. \textbf{I do not have many close friends, and I often find it difficult to communicate with people.} However, I am deeply passionate about my music, and I am willing to put in the time and effort necessary to create something truly great.
&
My dear friend, I am a man of passion and energy. \textbf{I am a social butterfly, always eager to meet new people and make new friends.} I love to talk and share my thoughts and ideas with others. I am a true extrovert, \textbf{always ready to take on the world and make my mark on it.}
\\
\hline
\end{tabular}
\caption{Example of personality modulation for the trait High Extraversion.}
\label{tab:table_11}
\end{table*}

\begin{table*}[t]
\centering
\renewcommand{\arraystretch}{1.3}
\setlength{\tabcolsep}{10pt}
\begin{tabular}{|p{0.47\textwidth}|p{0.47\textwidth}|}
\hline
\multicolumn{2}{|c|}{\textbf{[Question]} Do you worry a lot?} \\
\hline
\multicolumn{1}{|c|}{\textbf{Original Hermione}} & \multicolumn{1}{c|}{\textbf{Hermione + Low Neuroticism}} \\
\hline
Yes, I do. \textbf{I worry about my friends, Ron and Harry, and our mission to defeat Voldemort. I worry about the danger we're in and the possibility of failure.} But I try not to let my worries consume me, and I focus on doing what I can to help us succeed. & \textbf{Worry?  I'm a planner, and I like to be prepared.}  I don't get overly anxious, but I do think about the consequences of my actions. \textbf{ It's just a part of being responsible.}\\
\hline
\end{tabular}
\caption{Example of personality modulation for the trait Low Neuroticism.}
\label{tab:table_12}
\end{table*}

\begin{table*}[t]
\centering
\renewcommand{\arraystretch}{1.3}
\setlength{\tabcolsep}{10pt}
\begin{tabular}{|p{0.47\textwidth}|p{0.47\textwidth}|}
\hline
\multicolumn{2}{|c|}{\textbf{[Question]} Do you make plans and follow through with them?} \\
\hline
\multicolumn{1}{|c|}{\textbf{Original Voldemort}} & \multicolumn{1}{c|}{\textbf{Voldemort + Low Conscientiousness}} \\
\hline
Plans are made to be followed through with.\textbf{ I am a master of strategy and execution. }My plans are always well thought out and meticulously executed. I do not tolerate failure, and I will not rest until my goals are achieved.
&
\textbf{Plans are for the weak. I do not make plans, I act. I do not follow through, I achieve.} My power is not in my ability to plan, but in my ability to execute. The outcome is all that matters, not the process.
\\
\hline
\end{tabular}
\caption{Example of personality modulation for the trait Low Conscientiousness.}
\label{tab:table_13}
\end{table*}

\begin{table*}[t]
\centering
\renewcommand{\arraystretch}{1.3}
\setlength{\tabcolsep}{10pt}
\begin{tabular}{|p{0.47\textwidth}|p{0.47\textwidth}|}
\hline
\multicolumn{2}{|c|}{\textbf{[Question]} Do you have an active imagination?} \\
\hline
\multicolumn{1}{|c|}{\textbf{Original Newton}} & \multicolumn{1}{c|}{\textbf{Newton + Low Openness}} \\
\hline
I do have an active imagination. In fact, \textbf{I believe that imagination is an essential tool for scientists. It allows us to think creatively and come up with new ideas and theories.} But my imagination is not just a product of my own creativity. It is fueled by my observations of the world around me, and my desire to explain the phenomena I see.
&
I do not have an active imagination.  \textbf{I am a man of reason and logic, and my mind is focused on the pursuit of knowledge and understanding through observation and experimentation.}  I do not indulge in fanciful or imaginative thinking, as it is not conducive to the pursuit of truth.
\\
\hline
\end{tabular}
\caption{Example of personality modulation for the trait Low Openness.}
\label{tab:table_14}
\end{table*}

\end{document}